# Genetic Transfer or Population Diversification? Deciphering the Secret Ingredients of Evolutionary Multitask Optimization

Abhishek Gupta and Yew-Soon Ong

School of Computer Science and Engineering, Nanyang Technological University, Singapore
Email: {abhishekg, asysong}@ntu.edu.sg

*Abstract* — Evolutionary multitasking has recently emerged as a novel paradigm that enables the similarities and/or latent complementarities (if present) between distinct optimization tasks to be exploited in an autonomous manner simply by solving them together with a unified solution representation scheme. An important matter underpinning future algorithmic advancements is to develop a better understanding of the driving force behind successful multitask problem-solving. In this regard, two (seemingly disparate) ideas have been put forward, namely, (a) *implicit genetic transfer* as the key ingredient facilitating the exchange of high-quality genetic material across tasks, and (b) *population diversification* resulting in effective global search of the unified search space encompassing all tasks. In this paper, we present some empirical results that provide a clearer picture of the relationship between the two aforementioned propositions. For the numerical experiments we make use of Sudoku puzzles as case studies, mainly because of their feature that outwardly unlike puzzle statements can often have nearly identical final solutions. The experiments reveal that while on many occasions "genetic transfer" and "population diversity" may be viewed as two sides of the same coin, the wider implication of genetic transfer, as shall be shown herein, captures the true essence of evolutionary multitasking to the fullest.

## I. Introduction

In today's world of rapidly increasing volume, speed, and complexity of real-world challenges, the ability to effectively multitask, both cognitively as well as on computational platforms, is gaining much importance with the potential for *productivity enhancement* acting as the key motivation [1]. While some notable success stories can be found in the field of machine learning [2], much is yet to be explored in the context of numerical optimization and the biologically inspired search algorithms of computational intelligence. In this regard, recent works towards enhancing the population-based search algorithms of evolutionary computation (EC) for tackling multiple optimization tasks at once with a unified representation scheme, have shown considerable promise [3], [4]. In fact, it is noted that the implicit parallelism of a population is naturally well suited for the purpose of multitasking, providing the scope for spontaneous online sharing of knowledge building-blocks (in the form of encoded genetic material) across tasks.

In order to further emphasize the practical significance of the above, we consider the *extreme case* of two minimization tasks $T_1$ and $T_2$ characterized by cost functions $f_1$ and $f_2$ of *high ordinal correlation* [5]. In other words, for any pair of points $y_1$ and $y_2$ in a unified search space $Y$, $f_1(y_1) < f_1(y_2) \Leftrightarrow f_2(y_1) < f_2(y_2)$. Thus, on solving the two tasks together via multitasking, any series of search steps leading to a cost reduction of $T_1$ automatically leads to a cost reduction of $T_2$ *for free*, and vice versa, without the need for added function evaluations. As a result, it is clear that at least within the family of functions with high ordinal correlation, the idea of evolutionary multitasking (or multitask optimization) gives rise to *free lunches* [6], [7].

Recently in [8], a novel *multifactorial evolutionary algorithm* (MFEA) has been proposed as a means of exploiting the relationship between optimization tasks via the process of multitasking. The algorithm's nomenclature follows from the fact that each task is viewed as a unique factor influencing the evolution of a single population of individuals (artificial search agents). The algorithm provides a *cross-domain optimization platform* and has been tested on a variety of benchmarks as well as real-world instances, involving continuous and discrete problems, often leading to noteworthy results (a summary can be found in [1]).

The effectiveness of evolutionary multitasking showcased so far in the preliminary development stage has raised some interesting questions that are expected to direct future algorithmic advancements. In particular, it is considered critical to decipher the key ingredients driving the success of evolutionary multitasking. In this regard, two seemingly unrelated ideas have been put forward, namely, (a) implicit transfer of high-quality genetic building-blocks across tasks, and (b) population diversity leading to better global search.

It is noted that the boon of population diversification has been strongly emphasized in the design of island models of parallel genetic algorithms for single-tasking [9], [10], which may be seen to have some connections with the multitasking MFEA due to the implicit creation of subpopulations [8]. As a result, there is a tendency to assign the credit of MFEA's success merely to the exploitation of population diversity. However, doing so is contended to not adequately capture the true essence of evolutionary multitasking, or at least to be too premature. For starters, the advantage of population diversification is mainly reaped by subsequent genetic migrations (or transfer) across spatially distributed gene pools in complex search spaces [10]. In other words, the effectiveness of population diversity is itself reliant on the occurrence of genetic transfer. Moreover, in this paper, we shall empirically show that in addition to complementing population diversity, genetic transfer across tasks often plays the counter role of rapidly converging or streamlining the search towards promising regions of the search space. Thus, taking all observations into account, the notion of genetic transfer is contended to be a more appropriate metaphor for describing the mechanisms of evolutionary multitasking, one that captures its essential facets to the fullest.

For our empirical demonstrations, we make use of Sudoku puzzles as case studies, primarily due to two interesting features, namely, (a) they can be modelled as optimization problems [11]-[13] and (b) a pair of outwardly unlike puzzle statements may have nearly the same final solution. The second feature acts as an intuitive analogy for certain real-world problem-solving tasks which although appear different on the surface, may possess some latent similarities and/or complementarities. It must however be kept in mind that the purpose of this paper is not to propose a new algorithm for Sudoku puzzles, but to use them as a means to study the phenomena of genetic transfer and population diversification in evolutionary multitasking. The extent of the latter phenomenon is in fact explicitly computed during the evolutionary search via an *entropy* measure adopted from *information theory* [14], [15].

In order to provide a thorough exposition of the topic discussed heretofore, the remainder of the paper is organized as follows. Section II contains the preliminaries and a brief overview of the MFEA. Section III presents the general idea of Sudoku puzzles, their formulation as optimization tasks, and the associated genetic operators incorporated in the MFEA. Next, Section IV highlights the computational revelations of our study. Finally, some concluding remarks and a summary of the work are presented in Section V.

## II. THE MULTIFACTORIAL EVOLUTIONARY ALGORITHM

In this section, we first present the preliminaries with definitions of certain terms used to describe the MFEA. Thereafter, a brief overview of the algorithm is provided. For a more complete discussion on the bio-cultural motivation behind the algorithm, its roots in memetic computation [16], the importance of describing an appropriate unified representation scheme, etc., the reader is referred to [8].

### A. Preliminaries

Consider the case where $K$ optimization tasks are being tackled at the same time, within a single solver. With the ultimate goal of effective evolutionary multitasking, it is important to formulate a standard approach for comparing the fitness of candidate solutions associated with different tasks. In order to achieve this, we define a pair of properties describing every individual $p_i$, where $i \in \{1, 2, …, |P|\}$, in a population $P$. Note that every individual is encoded into a unified space $Y$, and can be translated into a task-specific solution with respect to any of the $K$ optimization tasks.

**Definition 1** (*Skill Factor*): The skill factor $\tau_i$ of $p_i$ is the one task, amongst all other tasks in a $K$-factorial environment, with which the individual is associated. If $p_i$ is evaluated for all tasks then $\tau_i = argmin_j\{r_j^i\}$, where $j \in \{1, 2, .., K\}$.

**Definition 2** (*Scalar Fitness*): The scalar fitness of $p_i$ in a multitasking environment is given by $\varphi_i = 1/r_{\tau_i}^i$.

### B. Algorithm Description

The pseudocode of the MFEA is provided in Algorithm 1. The procedure begins by generating a random initial population of individuals, denoted as $P_0$, in a unified search space. In subsequent generations, $P_t$ denotes the current population and $C_t$ denotes the child population. Notice that the pseudocode essentially follows the structure of a standard evolutionary algorithm, with the inclusion of *skill factor* and *scalar fitness* to accommodate for multiple optimization tasks at once.

---

**Algorithm 1:** Pseudocode of the MFEA

---
1. Generate $N$ random initial population $P_0$
2. **for every** $p_i$ in $P_0$ **do**
       Randomly assign a *skill factor* $\tau_i$
       Evaluate $p_i$ for task $\tau_i$ only
3. **end for**
4. Compute *scalar fitness* $\varphi_i$ for every $p_i$
5. Set $t = 0$
6. **while** (stopping conditions are not satisfied) **do**
       $C_t$ = Crossover+Mutate($P_t$)
       **for every** $c_i$ in $C_t$ **do**
           Determine skill factor $\tau_i$ via *imitation*
           Evaluate $c_i$ for task $\tau_i$ only
       **end**
       $R_t = C_t \cup P_t$
       Update *scalar fitness* of all individuals in $R_t$
       Select $N$ fittest members from $R_t$ to form $P_{t+1}$
       Set $t = t + 1$
7. **end while**

---

As noted earlier, the skill factor in Algorithm 1 represents the one task, amongst all other tasks in the multitasking environment, with which a particular individual in the population is associated. While skill factors are randomly assigned in the initial population, their propagation to subsequent offspring populations follows the principle of vertical cultural transmission [17], according to which a child *imitates* the skill factor of any one of its parents (at random). Thus, in an indirect way, the assignment of skill factors causes the overall population to be implicitly split into different subpopulations catering to different tasks. *Interestingly, it is possible that the genetic material created within one subpopulation (associated with a particular task) is also good for another. The opportunity for genetic transfer across tasks arises automatically in the MFEA when individuals possessing different skill factors crossover* [3]. The convenient feature of evolution is that whenever the transfer is positive (i.e., useful), the transferred genetic material is preserved following the principles of *natural selection*. On the other hand, if the transfer is negative (i.e., harmful), the same evolutionary principle kicks in to eliminate individuals carrying the deleterious genetic material.

Fig. 1. Statements of the six Sudoku puzzles considered in this paper (drawn from websudoku.com with some hand-crafted variations). Notice that all puzzles appear to be distinct on the surface. However, A1 and A2 have final solutions that are alike, while A1 and A3 are indeed different. Similarly, B1 and B2 have final solutions that are alike, while B1 and B3 are different.

## III. SOLVING SUDOKU PUZZLES WITH EC

There have been a number of works in recent years that have focused on solving hard Sudoku puzzles [11]-[13]. Accordingly, we reiterate that the purpose of this work is not to propose a new algorithm for Sudoku puzzles. Instead, we intend to utilize some of the features of the puzzle as a means to better understand the relationship between "genetic transfer" and "population diversification". The main feature of interest is that a pair of puzzles which appear to be completely different on the surface can possibly end up having final solutions that are alike, thereby providing an analogy to the prevalence of underlying synergies between seemingly distinct real-world problem-solving tasks.

A Sudoku puzzle is set on a 9 x 9 grid, where the aim is to fill every row, column, and each of nine 3 x 3 subgrids with digits from 1 to 9. The puzzle statement specifies a partially completed grid which most often has a unique final solution. As is clear, the initially filled squares are assumed to be fixed positions and their values cannot be altered while solving the puzzle. In addition to this, a Sudoku puzzle presents three more constraints to be satisfied: (a) the rows are permutations without digit repetition, (b) the columns are permutations without digit repetition, and (c) each of the nine 3 x 3 subgrids are permutations without digit repetition. A depiction of the partially filled grids for all six puzzles considered in this paper are provided in Fig. 1.

In [11], it has been shown that Sudoku can be cast as a combinatorial optimization problem by treating some of the constraints as hard constrains, and the remaining as soft constraints that can only be partially fulfilled, with the level of fulfillment representing the objective value to be maximized.

In this paper, we treat the row permutations as hard constraints that must be satisfied, while the column and subgrid permutations are treated as the soft constraints. Accordingly, the final fitness function to be maximized is equal to the number of unique elements in each column and each of the nine 3 x 3 subgrids. Thus, the maximum possible objective function value is given by $9 \cdot 9 + 9 \cdot 9 = 162$.

For satisfying the row permutation constraint, we incorporate the well-known partially matched crossover (PMX) [18] operator into the MFEA. The PMX is applied in a row-wise manner, as has been proposed in [11]. Notice that once the initial population in an evolutionary algorithm accounts for the specified fixed positions, the PMX operator ensures preservation of the fixed positions in subsequent generations. Finally, with regard to the mutation operator, two randomly chosen non-fixed positions in a row are exchanged, given some user-defined mutation probability.

## IV. COMPUTATIONAL REVELATIONS

In this section, we first provide the setup of the numerical experiments. Then, the entropy-based population diversity measure is described. Thereafter, the computational results and associated discussions are presented highlighting the observed relationship between "genetic transfer" and "population diversification".

### A. Experimental Setup

The initial problem statements of six Sudoku puzzles considered in this paper are depicted in Fig. 1. These puzzles have been drawn from the websudoku.com website, with some hand-crafted variations.

In the computational study, we carry out experiments for a variety of multitasking instances, and compare the obtained performance characteristics against single-tasking. For fairness of comparison, both approaches (i.e., single-tasking as well as multitasking) employ identical encoding scheme, population size, crossover operator (PMX), mutation operator (random swap), and other parameter settings.

To elaborate, each individual in the evolutionary algorithms is encoded by a 9 x 9 matrix, with each row being a permutation of the digits from 1 to 9. Note that since all tasks considered herein belong to the same domain, the need for describing an appropriate unification scheme for the MFEA is automatically bypassed. All tasks can directly be evaluated based on the aforementioned representation. However, the process of multitasking does introduce a challenge with regard to preservation of the fixed positions (see Section III), particularly when individuals possessing different skill factors (i.e., associated with different Sudoku puzzles) undergo recombination. The supposed feature of the PMX operator to preserve the fixed positions no longer holds true in the case of multitasking as the fixed positions for different Sudoku puzzles are different. In order to tackle this issue, once an offspring has been assigned a skill factor by the process of imitation (see Algorithm 1), it undergoes a repair step to ensure that all the fixed position constraints of its associated Sudoku puzzle are satisfied.

Further, a population of 500 individuals is deployed with mutation probability of 0.8 during single- and multitasking.

We carry out a total of eight sets of experiments with different combinations of Sudoku puzzles that have been chosen with no particular preference; these are depicted and described in Fig. 1. For ease of experimental investigation, the puzzles have been arbitrarily partitioned into two groups, namely, **A** and **B**, mainly ensuring that each group allows for the following experimental setups: (a) a single-tasking problem, (b) a multitasking instance where two tasks are identical *clones*, (c) a multitasking instance where two tasks are superficially distinct but are known to have underlying synergies, and finally, (d) a multitasking instance where the tasks are completely different. For each setup 20 independent runs are performed with averaged results provided throughout.

### B. Entropy-based Diversity Measurement

The main purpose of this paper is to clearly bring out the relationship between the concepts of "genetic transfer" and "population diversification" as they pertain to evolutionary multitasking. A well-known method for quantifying the diversity of a population during various stages of evolution is based on the entropy measure in information theory [14].

For the representation scheme employed in this paper, we first compute the diversity at each locus, which is denoted as $E_{(r,s)}$ for the $r^{th}$ row and the $s^{th}$ column of the matrix (see illustration in Fig. 2), as follows,

$$E_{(r,s)} = -\sum_{j=1}^{9} pr(allele_{(r,s)} = j) \cdot \log_9 [pr(allele_{(r,s)} = j)]. \quad (1)$$

Here, $allele_{(r,s)}$ represents the digits contained in the locus $(r,s)$ of the matrix encoding. Further, $pr(z)$ is the probability of '$z$' occurring in the entire population at a given generation. Notice that $E_{(r,s)}$ can take a minimum value of 0 and maximum value of 1. Accordingly, the entropy of the entire population, summed over all loci of the matrix encoding, is given by,

$$E = \frac{1}{81} \sum_{r=1}^{9} \sum_{s=1}^{9} E_{(r,s)}. \quad (2)$$

Once again, due to the factor introduced outside the summation, the total entropy $E$ is bounded between 0 and 1. As is perhaps clear, a larger value of $E$ indicates greater diversity in the population.

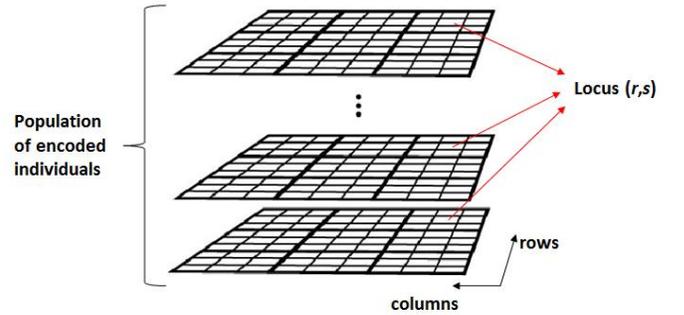

Fig. 2. Population of encoded individuals for which the diversity is first calculated at each locus using the entropy measure given by Eq. (1).

### C. Numerical Results and Discussions

For clarity of exposition, we present the results for each group of puzzles separately, emphasizing on the convergence characteristics achieved by the single- and multitasking approaches, as well as the evolution of population diversity.

*1) Group A puzzles*

A summary of the convergence characteristics achieved for various combinations of puzzles within group **A** is presented in Fig. 3. We recall that the puzzles A1 and A2, albeit appearing different on the surface, are known to possess underlying synergies. On the other hand, puzzles A1 and A3 are completely different to the best of our knowledge. Accordingly, based on the fundamental motivation behind evolutionary multitasking that sufficiently correlated optimization tasks can be solved faster when bundled together, it is not surprising to find that the coupling of A1 and A2 leads to the most superior convergence characteristics. Interestingly, the combination of A1 with its clone leads to significantly worse performance (see Fig. 3), implying that the observed complementarity between optimization tasks stems from features that may be different from mere task similarity. Even when completely dissimilar tasks A1 and A3 are solved together, the convergence behavior during the initial stages of the evolutionary search process (i.e., approximately the first 1.8e4 function evaluations) is visibly better compared to single-tasking as well as multitasking across identical clones.

Regarding the combination of A1 with itself, the MFEA has the effect of implicitly splitting the population into two islands (or subpopulations) both of which attempt to solve the exact same task. As a result, it is possible, as observed here, that the genetic lineage in both subpopulations progress in tandem with no mutually beneficial information. *However, when multiple (possibly diverse) tasks are executed simultaneously, the genetic transfer across them can bring about substantial leaps in the genetic lineage every now and then, thereby accelerating convergence characteristics* [19].

genetic transfer being predominantly negative. Note that the population of individuals can still be driven to the global optimum for puzzle A1 simply by using a larger population size to combat the negative transfer. In this paper, we have used a significantly smaller population than other works on solving Sudoku puzzles [11] mainly to highlight the threat of negative transfer and the need to effectively overcome it during multitasking.

For the combination of A1 and A2, the total entropy of the population is found to rapidly drop to 0 (as shown in Fig. 4), indicating speedy convergence towards the global optimum. Thus, in this case, *the genetic transfer during evolutionary multitasking is found to harness the underlying synergy between the tasks, promptly streamlining the search towards the most promising regions of the search space* (*an effect that is notably opposite to that of population diversification*). Clearly, the genetic transfer occurring from task A2 to task A1 (and vice versa) is positive, highlighting the efficacy of multitasking when complementary tasks are solved together.

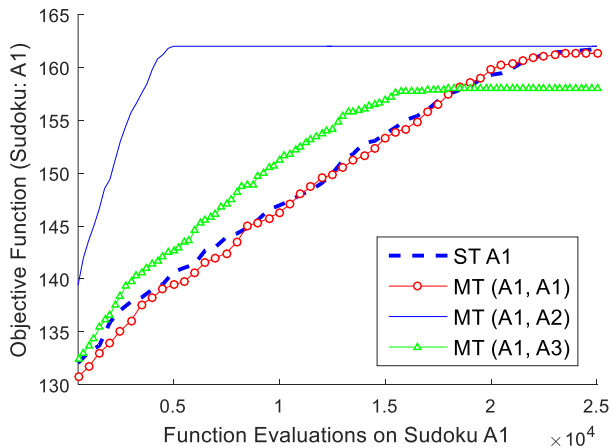

Fig. 3. Convergence characteristics for various single- and multitasking settings within group **A** puzzles. ST: single-tasking; MT: multitasking.

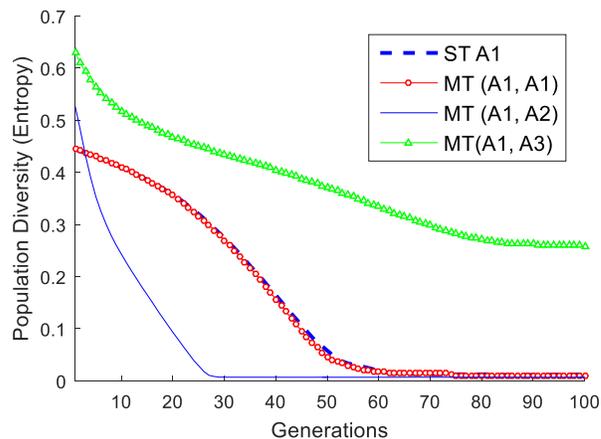

Fig. 4 Population diversity evolution trends for various single- and multitasking settings within group **A** puzzles. ST: single-tasking; MT: multitasking.

In order to fully appreciate the causes behind the observations in Fig. 3, we refer to the population diversity evolution trends in Fig. 4. For the case of combining unrelated tasks A1 and A3, it can be seen that the population diversity is consistently very high. Accordingly, during the initial stages of the evolutionary process, *the large population diversity coupled with the occurrence of genetic transfer across diverse tasks leads to improved global search*. In particular, within this period, the transfer of genetic material from task A3 to task A1 can be regarded as a form of *positive transfer* [20]-[22]. The aforesaid readily explains the observation in Fig. 3 where the convergence characteristics for MT (A1, A3) is found to be superior to single-tasking as well as multitasking across clones, during the initial stages of evolution. However, during the latter stages, the excessive diversity impedes convergence to the global optimum for MT (A1, A3), with the

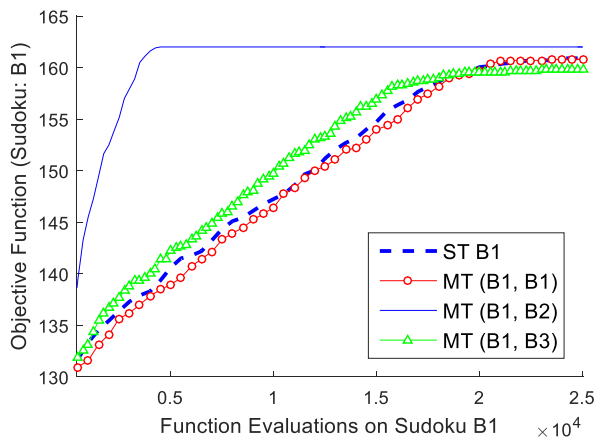

Fig. 5. Convergence characteristics for various single- and multitasking settings within group **B** puzzles. ST: single-tasking; MT: multitasking.

### 2) Group B puzzles

For the combination of puzzles in group **B**, we find qualitatively similar trends in the results as compared to the group **A** puzzles. This is evidenced by Figs. 5 and 6. *Once again, multitasking across distinct tasks with underlying synergies, namely, B1 and B2, leads to the best performance.* The spontaneous convergence of the population is demonstrated by the rapid drop in population diversity, as shown in Fig. 6, for MT (B1, B2). *In other words, the claim that genetic transfer can often have the effect of promptly streamlining search towards promising regions of the search space, is further reinforced.*

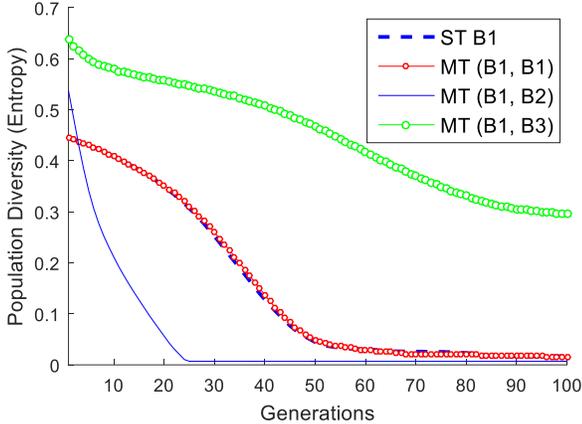

Fig. 6 Population diversity evolution trends for various single- and multitasking settings within group **B** puzzles. ST: single-tasking; MT: multitasking.

### 3) Summary of experiments

To summarize the computational experiments, the main message to be drawn from the results is that implicit genetic transfer is undoubtedly the key ingredient driving the performance of the evolutionary multitasking engine. *On the one hand, it is shown to complement population diversification when arbitrary tasks are put together in a single multitasking environment, thereby enhancing global search capabilities and accelerating convergence during the initial stages of evolution. On the other hand, when the tasks do possess underlying complementarities, genetic transfer kicks in to abandon diversification and rapidly focus the search onto the most promising regions. Notably, it is the facilitation of genetic transfer across tasks (not population diversification) that acts as the single constant across the different scenarios.*

Finally, we also highlight the observation that highly similar tasks, mimicked by cloning tasks in the present paper, may not necessarily be mutually complementing. In other words, *it is contended that complementarity between tasks may in fact be a feature distinct from mere task similarity*.

### D. Genetic Transfer and Task Complementarity

Roughly speaking, while solving a pair of tasks together in a unified solution representation space $Y$, the complementarity of task $T_2$ towards task $T_1$, given a specified set of genetic operators, may be explained as follows:

*For traversing from a solution $y_1$ to a more preferred solution $y_2$ with respect to task $T_1$, if using the fitness landscape of $T_2$ is more favorable than using the fitness landscape of $T_1$ alone (given available genetic operators), then $T_2$ is said to complement $T_1$ at least locally.*

For the case of Sudoku, we analyze the aforesaid by considering two puzzles that have different fixed positions but final solutions that are alike. Let the number of fixed positions for $T_1$ be $fp_1$ and the number of fixed positions for $T_2$ be $fp_2$. The total number of possible candidate solutions for $T_1$ in a naïve evolutionary solver (i.e., without problem-specific knowledge incorporation) is given by $9^{(81 - fp_1)}$, of which only one is globally optimum. On the other hand, the number of possible solutions for $T_1$ conditioned on the fixed positions of $T_2$ is approximately $9^{(81 - fp_1 - fp_2)}$. Since $9^{(81 - fp_1 - fp_2)} < 9^{(81 - fp_1)}$, the presence of $T_2$ clearly complements $T_1$ during multitasking by effectively shrinking the size of the search space. Interestingly, during task cloning, the size of the search space remains $9^{(81 - fp_1)}$, thereby not showing any complementarity.

In combinatorial domains, changes in the probability distribution over the set of possible solutions to a particular task, conditioned upon the transfer of genetic building-blocks (schema [23]) from other tasks, can be traced. However, a similar analysis in the domain of real-parameter optimization is considerably more challenging. A recent work that formulates a *functional synergy metric* (FSM) to explicitly compute the complementarity between continuous objective functions of distinct tasks can be found in [24].

## V. CONCLUSIONS

In this paper, we have presented an empirical study of the relationship between "genetic transfer" and "population diversification" as they pertain to the emerging topic of evolutionary multitasking. Using Sudoku puzzles as case studies, the results highlight that while on occasions genetic transfer and population diversity act as two sides of the same coin, the scope of genetic transfer extends to a larger variety of scenarios. In fact, for the particular case of multitasking across tasks with strong underlying complementarities, genetic transfer shows the effect of rapidly converging the population towards promising regions of the unified search space. Thus, we consider genetic transfer to be a more appropriate metaphor for explaining the mechanisms of multitask optimization, one that captures its true essence to the fullest.